\documentclass[conference]{IEEEtran}
\IEEEoverridecommandlockouts

\usepackage{cite}
\usepackage{amsmath,amssymb,amsfonts}
\usepackage{graphicx}
\usepackage{textcomp}
\usepackage{xcolor}
\usepackage{algorithm}
\usepackage{algpseudocode}
\usepackage{booktabs}
\usepackage{array}
\usepackage{adjustbox}
\usepackage{float}
\usepackage{tikz}
\usepackage{pgfplots}
\usepackage{hyperref}
\usepackage{cleveref}
\usepackage{hyperref}
\usetikzlibrary{positioning,arrows.meta}
\usetikzlibrary{patterns}

\def\BibTeX{{\rm B\kern-.05em{\sc i\kern-.025em b}\kern-.08em
    T\kern-.1667em\lower.7ex\hbox{E}\kern-.125emX}}

\begin{document}

\title{Embedding Graph Kernel Entropy (EGKE): Fine-grained Hallucination Detection for LLMs from Embeddings
\thanks{This work was supported by the French/US joint project LUCAS between the Agence Nationale de la Recherche (ANR) (grant ANR-25-CE23-2189) and the U.S. National Science Foundation (NSF) (grant numbers 2530255 and 2530256)}}


\author{
\IEEEauthorblockN{
Dilrangi Sankalpana\textsuperscript{1} \textsuperscript{\textsection},
Pragatheeswaran Vipulanandan\textsuperscript{2},
Uthayasanker Thayasivam\textsuperscript{1},
Kamal Premaratne\textsuperscript{2}
}

\IEEEauthorblockA{
\textsuperscript{1}Department of Computer Science and Engineering,
University of Moratuwa, Sri Lanka \\
\textsuperscript{2}Department of Electrical and Computer Engineering,
University of Miami, Coral Gables, Florida, USA \\
\textsuperscript{\textsection}dilrangi.21@cse.mrt.ac.lk
}
}

\maketitle

\begin{abstract}
Hallucination detection is critical for ensuring the reliability of Large Language Models (LLMs) in real-world deployments. While recent works, such as Semantic Entropy (SE) and Kernel Language Entropy (KLE), identify hallucinations by measuring semantic disagreement across multiple generated responses using Natural Language Inference (NLI) models, these methods depend on separately loaded NLI classifiers to estimate pairwise semantic similarity, introducing significant computational overhead. We propose Embedding Graph Kernel Entropy (EGKE), a variant of Kernel Language Entropy (KLE) that replaces its NLI-derived semantic graph with a Radial Basis Function (RBF) kernel applied to dense sentence embeddings generated by a BGE embedding model, while retaining KLE's graph Laplacian, heat kernel, and von Neumann entropy formulation unchanged. By computing similarities directly in a continuous embedding space, the proposed method provides a symmetric and computationally efficient alternative to NLI-based semantic comparison. We evaluate our approach across 30 experimental setups—covering multiple question-answer datasets and open-weight LLM families—on both AUROC and AUARC and demonstrate that this RBF-based method achieves competitive hallucination detection performance against state-of-the-art methods.
\end{abstract}


\section{Introduction}

Large Language Models (LLMs) have demonstrated remarkable performance in natural language processing; however, they remain prone to generating responses that are factually baseless, logically unsupported, or inconsistent with available knowledge. Such hallucinations remain a major obstacle to the deployment of LLMs in high-stakes domains such as healthcare, legal services, and education, where inaccurate outputs can lead to serious consequences. Consequently, developing reliable mechanisms for hallucination detection has become an important research challenge.

Hallucinations can arise in several forms. In this work, we focus specifically on \textit{confabulations}, a subtype of hallucination in which a model generates different plausible answers to the same question due to uncertainty about the correct response. Rather than producing a single confidently incorrect statement, a confabulating model exhibits semantic disagreement across multiple sampled outputs. Recent studies have shown that this disagreement can be quantified through semantic uncertainty, making it an effective signal for identifying confabulations in LLM-generated responses.

Most existing semantic uncertainty methods measure this disagreement by sampling multiple responses to the same question and quantifying how much their meanings diverge. Among these, Kernel Language Entropy (KLE) [6] estimates semantic uncertainty by constructing a semantic graph over sampled responses and computing the von Neumann entropy of a heat kernel derived from that graph's Laplacian. However, KLE relies on Natural Language Inference (NLI) models to assign the edge weights of this semantic graph, which introduces a second black box into the pipeline, requires a separate cross-encoder forward pass for every response pair, and offers limited transparency regarding why two responses are considered similar or different. We therefore propose Embedding Graph Kernel Entropy (EGKE), a variant of KLE in which the NLI-derived semantic graph is replaced by an RBF kernel over dense sentence embeddings, while the downstream graph Laplacian, heat kernel, and von Neumann entropy computation remain identical to KLE's formulation. This substitution makes similarity judgments explicit, symmetric, and computationally efficient, without altering the uncertainty-estimation machinery that KLE has already shown to be effective.

\section{Related Work}

Existing hallucination detection approaches can be broadly categorized including prompting-based methods such as SelfCheckGPT~\cite{manakul2023selfcheckgpt}, token probability and confidence-based approaches~\cite{kadavath2022language,quevedo2024detecting}, internal-state methods~\cite{azaria2023internal}, and semantic consistency methods based on natural language inference (NLI)~\cite{farquhar2024semantic,nikitin2024kle,kossen2024semantic,lei2023chain}. Among these, semantic approaches have emerged as an effective direction for detecting confabulations by comparing the semantic agreement between multiple generated responses.

\textit{Semantic Entropy (SE)}~\cite{farquhar2024semantic} introduced a semantic-level uncertainty measure for hallucination detection. Instead of treating lexical variation, SE groups generated responses into semantically equivalent clusters using bidirectional entailment, where two responses are considered equivalent only if each entails the other according to an NLI model. Semantic entropy is then computed over the resulting semantic clusters rather than over token sequences directly. This enables the method to distinguish genuine semantic disagreement from simple paraphrasing and improves hallucination detection compared to token-level entropy measures.

\textit{Kernel Language Entropy (KLE)}~\cite{nikitin2024kle} extends this idea by replacing discrete semantic clusters with a continuous similarity structure. Instead of assigning responses to strictly separate equivalence classes, KLE constructs a weighted semantic graph in which edge weights represent pairwise semantic similarity scores derived from NLI predictions. This addresses the ambiguity introduced by hard cluster membership in SE. A kernel matrix is then derived from the graph Laplacian, and semantic uncertainty is estimated using the von Neumann entropy of the resulting kernel. KLE demonstrates strong hallucination detection performance across multiple datasets and model families.

Other recent methods also leverage NLI-based semantics for hallucination detection. For example, SummaC~\cite{laban2022summac} applies an NLI model to verify factual consistency between source documents and generated summaries. Chen, Zhao, Karimi, and Flek introduce HaluMap~\cite{chen2025explainable}, a framework that detects hallucinations by mapping entailment and contradiction relations between source inputs and output segments, producing an interpretable hallucination heatmap. The CLATTER framework~\cite{eliav2025clatter}, which decomposes outputs into factual claims and performs stepwise entailment checks on each claim for fine-grained detection.

Along similar lines, the quantum tensor network framework of~\cite{vipulanandan2026semanticuncertaintyquantificationhallucinations} captures aleatoric uncertainty through semantic clustering. Additional approaches include Semantic Entropy Probes~\cite{kossen2024semantic} and Chain-of-NLI~\cite{lei2023chain}, which use lightweight or sequential NLI reasoning, as well as Semantic Nearest Neighbor Entropy (SNNE)~\cite{nguyen2025beyond} and Semantic Density~\cite{qiu2024semantic}, which incorporate pairwise semantic similarity between model outputs.

Within these methods, the NLI model performs a specific role: it maps a pair of generated responses to semantic relation labels such as entailment, contradiction, and neutral. These predictions are then converted into similarity scores used to construct semantic graphs or clusters. The remaining stages of the pipeline are numerical operations independent of the NLI model itself. This modular structure makes it possible to replace the NLI-based similarity computation stage without changing the downstream framework.

However, NLI-based approaches have practical limitations. They require a separate cross-encoder forward pass for every pair of responses, making the inference cost proportional to the number of response pairs. This is computationally expensive, as each call loads and runs a large classification model independently. Moreover, NLI classifiers are trained on coarse three-way classification tasks (e.g., MNLI)~\cite{williams2017broad}, so their outputs do not form a fine-grained similarity metric. Entailment is also inherently asymmetric: if response $A$ entails $B$, the reverse need not hold, requiring explicit symmetrization of scores. Embedding-based similarity avoids these issues: all responses are encoded in a single batched forward pass, pairwise distances are computed in closed form, and the resulting similarity scores are symmetric by construction and capture continuous degrees of semantic relatedness. This motivates replacing the NLI stage with direct embedding similarity in the uncertainty estimation pipeline.

\subsection{Our Contributions}

In this work, we propose an embedding-based hallucination detection approach (EGKE), a variant of KLE that models semantic similarity directly in a continuous embedding space using BAAI General Embedding(BGE). Pairwise similarities are computed using an RBF kernel, providing a symmetric and computationally efficient alternative to NLI-based semantic comparison.

The main contributions of this work are as follows:

\begin{itemize}
    
    \item We remove the dependency on NLI, such as DeBERTa,  based inference stage with an embedding-based semantic similarity mechanism using BGE representations and an RBF kernel. Unlike pairwise entailment inference, which requires a separate cross evaluation for each response pair and produces asymmetric relation scores, the proposed approach computes similarities directly in embedding space, yielding a symmetric similarity measure.

    \item We construct a fully connected semantic graph from embedding-based response similarities and derive a heat kernel representation over the graph. This provides a continuous and globally consistent characterization of semantic relationships between generated responses.
    
    \item Semantic uncertainty is estimated using von Neumann entropy computed over the resulting heat-kernel representation. This enables confabulation detection directly from semantic relationships encoded within the response graph.
    
    \item We evaluate the proposed approach against state-of-the-art semantic uncertainty methods using Area Under the Receiver Operating Characteristic curve (AUROC) and Area Under the Accuracy-Rejection Curve (AUARC) metrics across multiple QA benchmarks and LLM families.
\end{itemize}

\section{Methodology}

Given a question $q$, an LLM is prompted multiple times under stochastic decoding to generate a set of responses

\begin{equation}
S = {s_1, s_2, \ldots, s_n}.
\end{equation}

\begin{figure}[t]
\centering

\resizebox{\columnwidth}{!}{%
\begin{tikzpicture}[
    font=\Huge,
    >=Latex,
    node distance=0.9cm,
    arrow/.style={
        ->,
        thick,
    },
    bluebox/.style={
        draw=blue!80!black,
        rounded corners,
        thick,
        fill=blue!3,
        minimum width=11.0cm,
        minimum height=3.0cm,
        align=center
    },
    greenbox/.style={
        draw=teal!70!black,
        rounded corners,
        thick,
        fill=teal!5,
        minimum width=11.0cm,
        minimum height=4.0cm,
        align=center
    },
]
\node[bluebox] (question) {
{\Huge \textbf{Question + Context}}\\[1mm]
400 questions sampled per dataset
};
\node[bluebox, right=2.2cm of question] (llm) {
{\Huge \textbf{LLM response generation}}
};
\draw[arrow] (question) -- (llm);
\node[below=0.9cm of llm] (sampletext) {
10 high-temp samples ($T=1.0$)
};
\draw[arrow] (llm) -- (sampletext);
\node[greenbox, below=0.9cm of sampletext] (preprocess) {
{\Huge \textbf{Response preprocessing}}
};
\draw[arrow] (sampletext) -- (preprocess);
\node[greenbox, below=0.9cm of preprocess] (sbert) {
{\Huge \textbf{SBERT encoding}}\\[2mm]
Output: $n$ embedding vectors\\
$\mathbf{e}_1,\ldots,\mathbf{e}_n \in \mathbb{R}^{d}$
};
\draw[arrow] (preprocess) -- (sbert);
\node[greenbox, below=1.0cm of sbert] (rbf) {
{\Huge \textbf{RBF graph construction}}\\[2mm]
For each pair $(i,j)$:\\
$\mathbf{W}_{ij}
=
\exp\!\left(
-\frac{\|\mathbf{e}_i-\mathbf{e}_j\|^2}
{2\sigma^2}
\right)$
};
\draw[arrow] (sbert) -- (rbf);
\node[bluebox, below=0.9cm of rbf] (laplacian) {
{\Huge \textbf{Graph Laplacian}
}};
\draw[arrow] (rbf) -- (laplacian);
\node[
    bluebox,
    below=0.90cm of laplacian,
    minimum height=5.0cm,
] (heat) {
{\Huge \textbf{Heat kernel and kernel normalization}}\\[2mm]
$\mathbf{K}=\exp(-t\,\mathbf{L})$
\\[1.5mm]
$\hat{K}_{ij}
=
\dfrac{K_{ij}}
{\sqrt{K_{ii}K_{jj}}}
\cdot
\dfrac{1}{n}$
};
\draw[arrow] (laplacian) -- (heat);
\node[bluebox, right=2.6cm of heat] (entropy) {
{\Huge \textbf{Embedding Garph Kernel Entropy}}\\[2mm]
$\mathrm{EGKE}(\hat{\mathbf {K}})
=
-\sum_i \lambda_i \log \lambda_i$
\\[2mm]
(eigen decomposition)
};
\draw[arrow] (heat) -- (entropy);
\end{tikzpicture}%
}
\caption{Overview of the proposed Embedding Graph Kerenel Entropy pipeline. Responses generated by the LLM are embedded using BGE model, transformed into an RBF-weighted graph, and converted into a normalized heat kernel whose Von Neumann entropy is used as the semantic uncertainty estimate.}
\label{fig:rbf_kle_pipeline}

\end{figure}
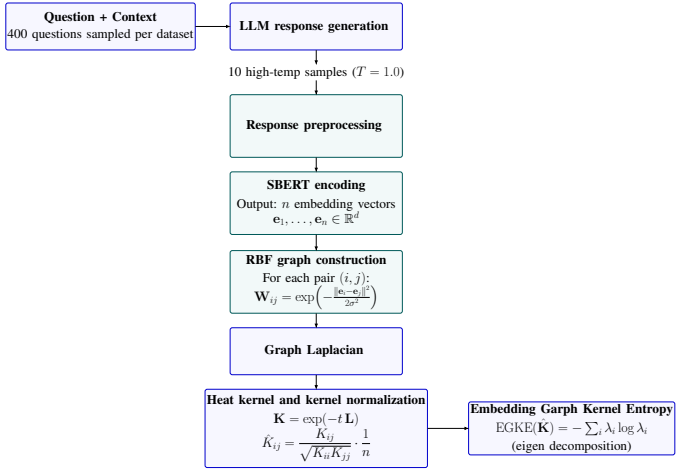

The objective is to estimate semantic uncertainty by measuring how consistently these responses occupy a shared semantic space. The proposed approach consists of four stages: response preprocessing, sentence embedding generation, RBF similarity graph construction, and heat kernel and von neumann entropy computation. An overview of the complete pipeline is shown in Fig.~\ref{fig:rbf_kle_pipeline}.

\subsection{Response Preprocessing}

Generated responses often differ in capitalization, punctuation, or other superficial lexical characteristics that do not alter their semantic meaning. To reduce the influence of such variations, each response is normalized by converting all text to lowercase and removing punctuation.

Since many question-answer datasets contain short answers whose meanings depend heavily on the associated question, the question is concatenated with the cleaned response before encoding:

\begin{equation}
\tilde{s}_i =
\texttt{"question: }q\texttt{ answer: }\mathrm{clean}(s_i)\texttt{"},
\end{equation}

where $\mathrm{clean}(\cdot)$ denotes lowercasing and punctuation removal. Prepending the question ensures that the encoder can resolve ambiguous short answers in context, which is important for QA tasks where responses such as entity names or numerical values carry little information in isolation.

\subsection{Sentence Embedding Generation}

Each preprocessed response $\tilde{s}_i$ is transformed into a dense semantic vector using the pretrained embedding model BAAI/bge-large-en-v1.5~\cite{xiao2023bge}. All responses corresponding to a question are encoded in a single batch, producing an embedding matrix

\begin{equation}
\mathbf{E} = 
[\mathbf{e}_1,\mathbf{e}_2,\ldots,\mathbf{e}_n]^{T}
\in \mathbb{R}^{n\times d},
\end{equation}

where $\mathbf{e}_i \in \mathbb{R}^{d}$ represents the embedding of response $s_i$ and $d$ is the embedding dimension.

These embeddings provide a continuous semantic representation in which geometrically nearby vectors correspond to semantically similar responses.

\subsection{RBF Similarity Graph Construction}

Given the embedding matrix, a fully connected weighted graph $G = (V, E)$ is constructed over the $n$ responses. Each response corresponds to a node, and every pair of nodes $(i, j)$ is connected by an edge with weight:
\begin{equation}
    w_{ij} = \exp\!\left(-\frac{\|\mathbf{e}_i - \mathbf{e}_j\|^2}{2\sigma^2}\right),
    \label{eq:rbf_main}
\end{equation}
where $\sigma > 0$ is the bandwidth hyperparameter. The squared Euclidean distance $\|\mathbf{e}_i - \mathbf{e}_j\|^2$ is computed directly from the embedding tensors, and the resulting weight lies in $(0, 1]$, attaining its maximum value of 1 when two responses are identical in embedding space and decaying toward 0 as they diverge. Unlike the NLI-based approach, which produces a sparse graph (edges only where entailment is detected), the RBF graph is fully connected, providing a smooth and globally consistent similarity structure over all response pairs.

\subsection{Embedding Graph Kernel Entropy (EGKE)}

Given the RBF-weighted graph $G$, the graph Laplacian is computed as $\mathbf{L} = \mathbf{D} - \mathbf{W}$, where $\mathbf{W}$ is the weight matrix with entries $W_{ij} = w_{ij}$ and $\mathbf{D}$ is the diagonal degree matrix with $D_{ii} = \sum_j W_{ij}$. The heat kernel is then derived as the matrix exponential of the scaled Laplacian: \begin{equation} K_t = e^{-t\mathbf{L}}, \label{eq:heat} \end{equation} $t$ is a diffusion length-scale parameter. Following KLE~\cite{nikitin2024kle}, we set t = 0.3, the reasonable default value reported by [6] and selected via their entropy convergence plots (Fig. 2 in~\cite{nikitin2024kle}); we did not perform a separate search over t and instead adopted this value unchanged to isolate the effect of the RBF-based graph construction from the choice of diffusion scale. The heat kernel $K_t$ is a positive-semidefinite matrix whose $(i, j)$ entry reflects the amount of information that diffuses between nodes $i$ and $j$ under a heat equation governed by the graph structure. Responses connected by high-weight edges--i.e., those with similar embeddings--are close in this diffusion metric and thus have large kernel values. 

The kernel is then normalised to unit trace via:

\begin{equation}
\hat{K}_{ij} = \frac{K_{ij}}
{\sqrt{K_{ii} \cdot K_{jj}}}
\cdot
\frac{1}{n},
\end{equation}

where $\hat{K}$ represents the semantic kernel constructed from the embedding graph.

We can now define \textbf{Embedding Graph Kernel Entropy (EGKE)} as the von Neumann entropy of this semantic kernel. Given a set of LLM generations ${S_1,\ldots,S_N}$ for an input question $x$, and a semantic kernel $\hat{K}$ derived from their embedding-based graph representation, EGKE is defined as

\begin{equation}
\mathrm{EGKE}(\hat{K}) = 
-\sum_i \lambda_i \log \lambda_i.
\label{eq:vne_main}
\end{equation}

A low EGKE value indicates that the semantic kernel has a few dominant eigenvalues, corresponding to a semantically concentrated response set with low semantic uncertainty. Conversely, a high EGKE value indicates a more uniformly distributed eigenspectrum, reflecting greater semantic diversity among generated responses and therefore higher semantic uncertainty. This scalar entropy value is reported for each question as the final semantic uncertainty estimate.

\section{Experiments}
 
\subsection{Experimental Setup}
 
\textbf{Datasets:} We evaluate on five question-answering benchmarks spanning distinct domains: TriviaQA~\cite{joshi2017triviaqa} (trivia knowledge), SQuAD~\cite{rajpurkar2016squad} (reading comprehension), Natural Questions (NQ)~\cite{kwiatkowski2019nq} (open-domain questions from real web queries), SVAMP~\cite{patel2021svamp} (elementary mathematical word problems), and BioASQ~\cite{tsatsaronis2015bioasq} (biomedical question answering). Context passages are withheld for all datasets except SVAMP, following the protocol of~\cite{nikitin2024kle}, to induce sufficient uncertainty for meaningful evaluation. For each dataset, 400 examples are used for evaluation.
 
\textbf{Models:} Experiments are conducted on six open-weight LLMs: Falcon-7B and Falcon-7B-Instruct~\cite{almazrouei2023falcon}, and LLaMA-2-7B and LLaMA-2-7B-Chat~\cite{touvron2023llama2}, and Mistral-7B-v0.1 and Mistral-7B-Instruct-v0.1~\cite{jiang2023mistral}. This selection includes both base and instruction-tuned variants, allowing us to assess whether the behaviour of the proposed method differs across training paradigms. For each question, $n = 10$ responses are sampled at temperature $T = 1.0$ using top-$K$ sampling ($K = 50$) and nucleus sampling ($p = 0.9$). An additional low-temperature sample ($T = 0.1$) is used to assess model accuracy against ground-truth answers.
 
\textbf{Baselines:} We compare our method against the following uncertainty estimation methods: Semantic Entropy (SE)~\cite{farquhar2024semantic}, Discrete Semantic Entropy (DSE)~\cite{farquhar2024semantic}, Predictive Entropy (PE)~\cite{malinin2020uncertainty}, P(True)~\cite{kadavath2022language}, and the Kernel Language Entropy with a heat kernel $\text{KLE}(K_\text{HEAT})$~\cite{nikitin2024kle}. SE and DSE represent the primary semantic clustering baselines. PE is the token-level entropy baseline. P(True) is a supervised in-context baseline that prompts the LLM to self-assess its own answer. $\text{KLE}(K_\text{HEAT})$ is a graph-based semantic uncertainty baseline that leverages pairwise semantic similarities between generated responses.
 
\textbf{Evaluation Metrics:} We report two metrics. The AUROC measures how well an uncertainty estimate discriminates between correct and incorrect model responses, treating the problem as binary classification. An AUROC of 1.0 indicates a perfect discriminator and 0.5 indicates chance-level performance. The AUARC~\cite{nadeem2009arc} measures the accuracy of the model on the subset of questions it is most confident about, summarised across all possible rejection thresholds. A higher AUARC indicates that the uncertainty estimate more reliably identifies questions where the model is likely to be correct.

\textbf{RBF Bandwidth Selection.} The proposed method introduces a single hyperparameter, the RBF bandwidth $\sigma$ which controls how rapidly the similarity between two responses decreases as the distance between their embeddings increases. Smaller values of $\sigma$ assign high similarity only to very closely related responses, whereas larger values allow even more distant responses to remain strongly connected. As the optimal bandwidth is dataset and model dependent, we evaluate a range of values $\sigma \in [0.1, 0.7]$ in parallel within each experimental run, with each value producing an independent graph and entropy estimate. We discuss the effect of this bandwidth choice and provide the corresponding sensitivity analysis in Appendix~\ref{app:hyperparameter_selection}.

\section{Discussion}

\Cref{fig:auroc_pairwise_winrate,fig:auarc_pairwise_winrate} summarize the pairwise win-rate comparison across all 30 model--dataset settings where each cell reports the fraction of settings in which the row method outperforms the column method, corresponding to six LLMs evaluated on five QA datasets. The complete numerical AUROC and AUARC results are reported in Appendix \ref{app:complete_results}, with model-wise results shown in Tables~\ref{tab:falcon-7b}--\ref{tab:mistral-7b-v0.1-instruct}. 
 
\vspace{0.3em}

\definecolor{rdBu0}{HTML}{67001F}
\definecolor{rdBu1}{HTML}{B2182B}
\definecolor{rdBu2}{HTML}{D6604D}
\definecolor{rdBu3}{HTML}{F4A582}
\definecolor{rdBu4}{HTML}{FDDBC7}
\definecolor{rdBu5}{HTML}{F7F7F7}
\definecolor{rdBu6}{HTML}{D1E5F0}
\definecolor{rdBu7}{HTML}{92C5DE}
\definecolor{rdBu8}{HTML}{4393C3}
\definecolor{rdBu9}{HTML}{2166AC}
\definecolor{rdBu10}{HTML}{053061}

\definecolor{diagblue}{HTML}{8BD3E6}

\newcommand{\CellColor}[1]{%
  \ifdim#1pt<0.15pt
    \xdef\cellfill{rdBu0}%
  \else\ifdim#1pt<0.25pt
    \xdef\cellfill{rdBu1}%
  \else\ifdim#1pt<0.35pt
    \xdef\cellfill{rdBu2}%
  \else\ifdim#1pt<0.45pt
    \xdef\cellfill{rdBu3}%
  \else\ifdim#1pt<0.50pt
    \xdef\cellfill{rdBu4}%
  \else\ifdim#1pt<0.55pt
    \xdef\cellfill{rdBu5}%
  \else\ifdim#1pt<0.65pt
    \xdef\cellfill{rdBu6}%
  \else\ifdim#1pt<0.75pt
    \xdef\cellfill{rdBu7}%
  \else\ifdim#1pt<0.85pt
    \xdef\cellfill{rdBu8}%
  \else\ifdim#1pt<0.95pt
    \xdef\cellfill{rdBu9}%
  \else
    \xdef\cellfill{rdBu10}%
  \fi\fi\fi\fi\fi\fi\fi\fi\fi\fi
}

\newcommand{\WinCell}[3]{%
  \CellColor{#3}%
  \draw[draw=white, line width=0.5pt, fill=\cellfill] (#1,#2) rectangle ++(1,1);
  \def\celltext{black}%
  \ifdim#3pt<0.35pt
    \def\celltext{white}%
  \fi
  \ifdim#3pt>0.72pt
    \def\celltext{white}%
  \fi
  \node[text=\celltext, font=\scriptsize] at (#1+0.5,#2+0.5) {#3};
}

\newcommand{\DiagCell}[2]{%
  \draw[draw=white, line width=0.5pt, fill=diagblue] (#1,#2) rectangle ++(1,1);
  \path[pattern=north east lines, pattern color=white] (#1,#2) rectangle ++(1,1);
  \path[pattern=north west lines, pattern color=white] (#1,#2) rectangle ++(1,1);
}

\newcommand{\HeatmapColorbar}{%
  \begin{scope}[shift={(7.55,0)}]
    \foreach \k in {0,...,100}{
      \pgfmathsetmacro{\v}{\k/100}
      \pgfmathsetmacro{\yone}{7*\k/100}
      \pgfmathsetmacro{\ytwo}{7*(\k+1)/100}
      \CellColor{\v}
      \draw[draw=\cellfill, fill=\cellfill] (0,\yone) rectangle (0.25,\ytwo);
    }
    \draw[black, line width=0.3pt] (0,0) rectangle (0.25,7);
    \foreach \v/\lab in {0.2/{0.2},0.4/{0.4},0.6/{0.6},0.8/{0.8}}{
      \node[anchor=west, font=\scriptsize] at (0.35,7*\v) {\lab};
    }
  \end{scope}
}

\newcommand{\HeatmapLabels}{%
  \foreach \i/\lab in {
    0/{EGKE(Ours)},
    1/{SE},
    2/{DSE},
    3/{KLE($K_{\mathrm{HEAT}}$)},
    4/{PE},
    5/{$P(\mathrm{True})$},
    6/{ER}
  }{
    \node[rotate=45, anchor=west, font=\scriptsize] at (\i+0.18,7.15) {\lab};
  }

  \foreach \i/\lab in {
    0/{EGKE(Ours)},
    1/{SE},
    2/{DSE},
    3/{KLE($K_{\mathrm{HEAT}}$)},
    4/{PE},
    5/{$P(\mathrm{True})$},
    6/{ER}
  }{
    \pgfmathtruncatemacro{\yy}{6-\i}
    \node[anchor=east, font=\scriptsize] at (-0.15,\yy+0.5) {\lab};
  }
}

\renewcommand{\HeatmapColorbar}[1][(7.55,0)]{%
  \begin{scope}[shift={#1}]
    \foreach \k in {0,...,100}{
      \pgfmathsetmacro{\v}{\k/100}
      \pgfmathsetmacro{\yone}{7*\k/100}
      \pgfmathsetmacro{\ytwo}{7*(\k+1)/100}
      \CellColor{\v}
      \draw[draw=\cellfill, fill=\cellfill] (0,\yone) rectangle (0.25,\ytwo);
    }
    \draw[black, line width=0.3pt] (0,0) rectangle (0.25,7);
    \foreach \v/\lab in {0.2/{0.2},0.4/{0.4},0.6/{0.6},0.8/{0.8}}{
      \node[anchor=west, font=\scriptsize] at (0.35,7*\v) {\lab};
    }
    \node[rotate=90, font=\scriptsize] at (0.95,3.5) {Win rate};
  \end{scope}
}


\begin{figure}[t]
\centering
\resizebox{\columnwidth}{!}{%
\begin{tikzpicture}

\DiagCell{0}{6}
\WinCell{1}{6}{0.50}
\WinCell{2}{6}{0.67}
\WinCell{3}{6}{0.82}
\WinCell{4}{6}{0.87}
\WinCell{5}{6}{1.00}
\WinCell{6}{6}{0.90}

\WinCell{0}{5}{0.50}
\DiagCell{1}{5}
\WinCell{2}{5}{0.57}
\WinCell{3}{5}{0.80}
\WinCell{4}{5}{0.88}
\WinCell{5}{5}{1.00}
\WinCell{6}{5}{0.97}

\WinCell{0}{4}{0.33}
\WinCell{1}{4}{0.43}
\DiagCell{2}{4}
\WinCell{3}{4}{0.78}
\WinCell{4}{4}{0.83}
\WinCell{5}{4}{1.00}
\WinCell{6}{4}{0.93}

\WinCell{0}{3}{0.18}
\WinCell{1}{3}{0.20}
\WinCell{2}{3}{0.22}
\DiagCell{3}{3}
\WinCell{4}{3}{0.65}
\WinCell{5}{3}{0.97}
\WinCell{6}{3}{0.87}

\WinCell{0}{2}{0.13}
\WinCell{1}{2}{0.12}
\WinCell{2}{2}{0.17}
\WinCell{3}{2}{0.35}
\DiagCell{4}{2}
\WinCell{5}{2}{0.97}
\WinCell{6}{2}{0.77}

\WinCell{0}{1}{0.00}
\WinCell{1}{1}{0.00}
\WinCell{2}{1}{0.00}
\WinCell{3}{1}{0.03}
\WinCell{4}{1}{0.03}
\DiagCell{5}{1}

\WinCell{6}{1}{0.37}

\WinCell{0}{0}{0.10}
\WinCell{1}{0}{0.03}
\WinCell{2}{0}{0.07}
\WinCell{3}{0}{0.13}
\WinCell{4}{0}{0.23}
\WinCell{5}{0}{0.63}
\DiagCell{6}{0}

\HeatmapLabels
\HeatmapColorbar

\end{tikzpicture}
}
\caption{Pairwise AUROC win-rate heatmap across 30 model--dataset settings. Each cell shows the fraction of settings in which the row method outperforms the column method.}
\label{fig:auroc_pairwise_winrate}
\end{figure}
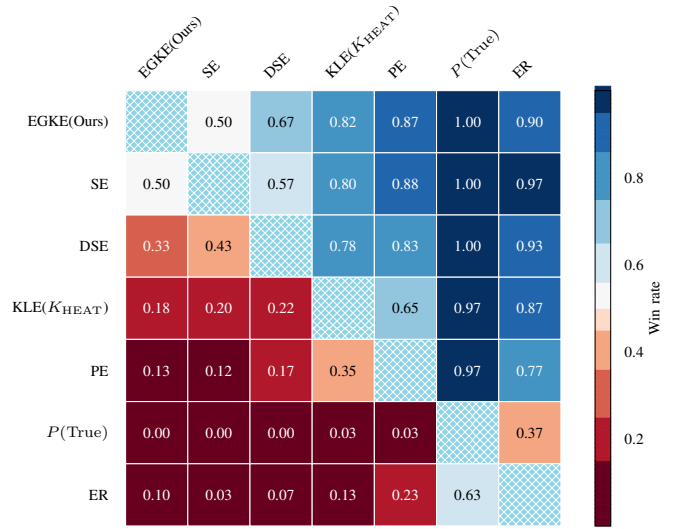


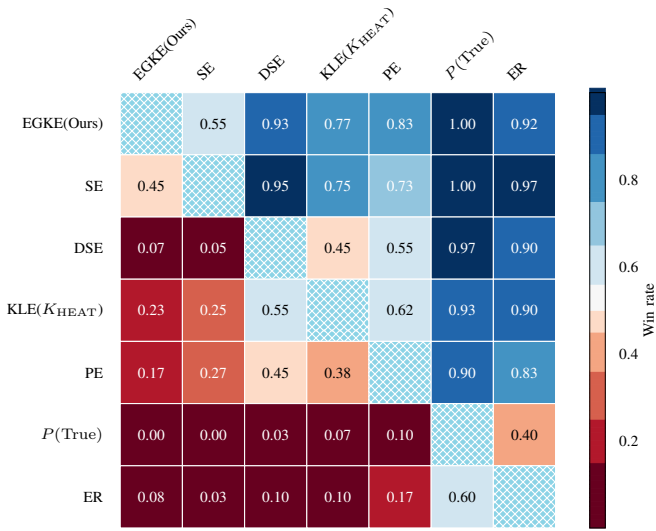
\begin{figure}[t]
\centering
\resizebox{\columnwidth}{!}{%
\begin{tikzpicture}

\DiagCell{0}{6}
\WinCell{1}{6}{0.55}
\WinCell{2}{6}{0.93}
\WinCell{3}{6}{0.77}
\WinCell{4}{6}{0.83}
\WinCell{5}{6}{1.00}
\WinCell{6}{6}{0.92}

\WinCell{0}{5}{0.45}
\DiagCell{1}{5}
\WinCell{2}{5}{0.95}
\WinCell{3}{5}{0.75}
\WinCell{4}{5}{0.73}
\WinCell{5}{5}{1.00}
\WinCell{6}{5}{0.97}

\WinCell{0}{4}{0.07}
\WinCell{1}{4}{0.05}
\DiagCell{2}{4}
\WinCell{3}{4}{0.45}
\WinCell{4}{4}{0.55}
\WinCell{5}{4}{0.97}
\WinCell{6}{4}{0.90}

\WinCell{0}{3}{0.23}
\WinCell{1}{3}{0.25}
\WinCell{2}{3}{0.55}
\DiagCell{3}{3}
\WinCell{4}{3}{0.62}
\WinCell{5}{3}{0.93}
\WinCell{6}{3}{0.90}

\WinCell{0}{2}{0.17}
\WinCell{1}{2}{0.27}
\WinCell{2}{2}{0.45}
\WinCell{3}{2}{0.38}
\DiagCell{4}{2}
\WinCell{5}{2}{0.90}
\WinCell{6}{2}{0.83}

\WinCell{0}{1}{0.00}
\WinCell{1}{1}{0.00}
\WinCell{2}{1}{0.03}
\WinCell{3}{1}{0.07}
\WinCell{4}{1}{0.10}
\DiagCell{5}{1}
\WinCell{6}{1}{0.40}

\WinCell{0}{0}{0.08}
\WinCell{1}{0}{0.03}
\WinCell{2}{0}{0.10}
\WinCell{3}{0}{0.10}
\WinCell{4}{0}{0.17}
\WinCell{5}{0}{0.60}
\DiagCell{6}{0}

\HeatmapLabels
\HeatmapColorbar

\end{tikzpicture}
}
\caption{Pairwise AUARC win-rate heatmap across 30 model--dataset settings. Each cell shows the fraction of settings in which the row method outperforms the column method; ties are counted as 0.5.}
\label{fig:auarc_pairwise_winrate}
\end{figure}

 
 
\textbf{Embedding-based similarity is a viable substitute for NLI.}
Our method achieves the highest average AUROC (0.793) and AUARC (0.553) across all 30 model--dataset settings, outperforming all compared baselines in aggregate. It also wins against most baselines in the majority of settings, with particularly strong gains over probabilistic baselines such as P(True) and ER, while remaining competitive with semantic baselines such as SE and DSE across datasets and models. These results suggest that replacing pairwise NLI inference with embedding-based similarity does not degrade hallucination detection performance.

More broadly, the pairwise win-rate analysis shows that our method is frequently tied with, or better than, strong semantic, probabilistic, and embedding-based uncertainty baselines. In particular, while its AUROC comparison with SE is largely tied, our method achieves a higher AUARC win rate, indicating stronger selective prediction behavior. Importantly, this is achieved without SE's DeBERTa-based NLI component. A symmetric, distance-based measure in embedding space is therefore sufficient to construct a semantic graph that captures meaningful disagreement among sampled responses, while avoiding the computational overhead of running a separately loaded cross-encoder over every response pair.
 
\textbf{The RBF bandwidth $\sigma = 0.1$ is consistently optimal.} The sharp decay of the RBF kernel at small $\sigma$ means that only embeddings that are very close in Euclidean distance receive high similarity scores. In the embedding space of \texttt{BAAI/bge-large-en-v1.5}, this corresponds to responses that are semantically near-identical. This behaviour is geometrically sensible for QA tasks, where a model that is uncertain tends to produce responses that are spread across a wide region of embedding space, while a confident model produces responses clustered tightly around a single point. A small $\sigma$ is therefore better able to distinguish these two regimes by assigning near-zero weight to all but the closest response pairs. As $\sigma$ increases, the similarity function becomes less selective, causing even semantically different responses to receive relatively high similarity scores. This results in a denser semantic graph where meaningful distinctions between responses are blurred, reducing the ability of the uncertainty measure to separate reliable generations from hallucinations. The consistent decline in AUROC and AUARC observed for larger $\sigma$ values supports this interpretation.

\textbf{Limitations.} The proposed method relies on the quality of the BGE embedding mode as a proxy for semantic similarity. For highly specialised domains such as biomedical QA, a general-purpose sentence encoder may not capture domain-specific semantic relationships as reliably as an NLI model fine-tuned on in-domain data. Additionally, the RBF kernel introduces a hyperparameter $\sigma$ that, while consistent across our experiments, may require tuning for new domains or model families. Unlike the entropy convergence plots proposed in~\cite{nikitin2024kle} for the heat kernel, no analogous principled selection criterion currently exists for the RBF bandwidth in this setting.

\section{Conclusion}
 
We proposed an embedding-based hallucination detection method, a variant of Kernel Language Entropy (KLE) ~\cite{nikitin2024kle} that replaces NLI-based pairwise similarity estimation with an RBF kernel over dense sentence embeddings. Rather than running a cross-encoder for every pair of sampled responses, all responses to a question are encoded in a single batched forward pass, and pairwise similarities are computed in closed form from Euclidean distances in embedding space. This produces a symmetric, continuously graded similarity structure that is fed directly into a graph Laplacian, heat kernel, and von Neumann entropy pipeline for hallucination scoring.
 
Evaluated across 30 experimental scenarios spanning five QA benchmarks and six open-weight LLMs, our method achieves the highest average AUROC and AUARC among all compared methods, outperforming the KLE in 24 of 30 scenarios on AUROC and 22 of 30 on AUARC, and dominating token-level and self-assessment baselines in the large majority of scenarios. Performance is consistent across both base and instruction-tuned models. 

Future work could explore domain-adapted sentence encoders for specialised QA settings and principled selection of the RBF bandwidth parameter from the geometric structure of the embedding space.

Code and instructions for reproducing our results are available at \url{https://github.com/Dilrangi-Sankalpana/Embedding-Graph-Kernel-Entropy}.

\bibliography{Dilrangi/ICKJ_2026}
\bibliographystyle{IEEEtran}

\clearpage
\onecolumn
\appendix

\subsection{\textbf{Hyperparameter Selection Details}}
\label{app:hyperparameter_selection}

To select the RBF bandwidth parameter $\sigma$, we performed a grid search over seven candidate values ranging from 0.1 to 0.7, with the heat kernel lengthscale fixed at t = 0.3 (following the default value used by KLE($K_{\mathrm{HEAT}}$)~\cite{nikitin2024kle}) throughout the search. For each value, experiments were conducted across all datasets and language models. Table~\ref{tab:hyperparameter_summary} reports the average AUROC obtained for each model, where the scores are averaged across all evaluated datasets. Across all six models, $\sigma = 0.1$ consistently achieved the highest average performance and was therefore selected for all subsequent experiments.

\begin{table}[H]
\centering
\caption{Summary of selected hyperparameters after grid search.}
\label{tab:hyperparameter_summary}
\small
\setlength{\tabcolsep}{6pt}
\renewcommand{\arraystretch}{1.1}
\begin{tabular}{ccccccc}
\toprule
$\boldsymbol{\sigma}$ & \textbf{falcon-7b} & \textbf{falcon-7b-instruct } & \textbf{llama-2-7b} & \textbf{llama-2-7b-chat} & \textbf{mistral-7b-v0.1} & \textbf{mistral-7b-v0.1-instruct}\\
\midrule
\textbf{0.1} & \textbf{0.770} &  \textbf{0.738} &  \textbf{0.814} &  \textbf{0.792} &  \textbf{0.819} &  \textbf{0.833}\\ 
0.2 & 0.732 &  0.714 &  0.788 &  0.779 &  0.795 &  0.808\\
0.3 & 0.716 &  0.704 &  0.776 &  0.775 &  0.782 &  0.795\\
0.4 & 0.710 &  0.700 &  0.772 &  0.773 &  0.776 &  0.790\\
0.5 & 0.707 &  0.698 &  0.770 &  0.772 &  0.773 &  0.788\\
0.6 & 0.706 &  0.697 &  0.769 &  0.772 &  0.771 &  0.787\\
0.7 & 0.705 &  0.697 &  0.768 &  0.771 &  0.771 &  0.786\\

\bottomrule
\end{tabular}
\end{table}

\subsection{\textbf{Complete Experimental Results}}
\label{app:complete_results}

This appendix provides the complete numerical results corresponding to the AUROC and AUARC win rate plots presented in the main text. Tables I-VI report the mean performance and standard deviation for each uncertainty estimation method across all model–dataset settings. These results include the proposed  method together with the baseline methods SE, DSE, KLE, PE, P (True), and ER. The best-performing method for each model–dataset setting is highlighted
in bold.

\begin{table}[H]
\centering
\caption{AUROC / AUARC on Falcon-7B across datasets. Best results per metric in \textbf{bold}.}
\label{tab:falcon-7b}
\scriptsize
\resizebox{\textwidth}{!}{%
\begin{tabular}{lcccccccccc}
\toprule
\textbf{Method} &
\multicolumn{2}{c}{\textbf{TriviaQA}} &
\multicolumn{2}{c}{\textbf{SQuAD}} &
\multicolumn{2}{c}{\textbf{NQ}} &
\multicolumn{2}{c}{\textbf{SVAMP}} &
\multicolumn{2}{c}{\textbf{BioASQ}} \\
\cmidrule(lr){2-3}\cmidrule(lr){4-5}\cmidrule(lr){6-7}\cmidrule(lr){8-9}\cmidrule(lr){10-11}
 & AUROC & AUARC & AUROC & AUARC & AUROC & AUARC & AUROC & AUARC & AUROC & AUARC \\
\midrule
SE 
& \textbf{0.779 $\pm$ 0.02} & 0.691 $\pm$ 0.03 
& 0.711 $\pm$ 0.03 & 0.344 $\pm$ 0.03 
& 0.731 $\pm$ 0.03 & 0.467 $\pm$ 0.03 
& 0.760 $\pm$ 0.03 & \textbf{0.467 $\pm$ 0.04} 
& \textbf{0.874 $\pm$ 0.02} & \textbf{0.655 $\pm$ 0.03}  \\

DSE
& 0.770 $\pm$ 0.02 & 0.682 $\pm$ 0.03 
& 0.698 $\pm$ 0.03 & 0.328 $\pm$ 0.03 
& 0.738 $\pm$ 0.03 & 0.454 $\pm$ 0.03 
& 0.738 $\pm$ 0.03 & 0.439 $\pm$ 0.04 
& 0.872 $\pm$ 0.02 & 0.636 $\pm$ 0.03 \\

PE
& 0.738 $\pm$ 0.03 & 0.673 $\pm$ 0.03
& \textbf{0.725 $\pm$ 0.03} & \textbf{0.351 $\pm$ 0.03}
& 0.751 $\pm$ 0.03 & 0.469 $\pm$ 0.03
& 0.740 $\pm$ 0.03 & 0.456 $\pm$ 0.04
& 0.804 $\pm$ 0.02 & 0.617 $\pm$ 0.03 \\

P\_true
& 0.576 $\pm$ 0.03 & 0.577 $\pm$ 0.03
& 0.546 $\pm$ 0.03 & 0.248 $\pm$ 0.03
& 0.566 $\pm$ 0.03 & 0.352 $\pm$ 0.03
& 0.560 $\pm$ 0.03 & 0.338 $\pm$ 0.03
& 0.714 $\pm$ 0.03 & 0.562 $\pm$ 0.03 \\

ER
& 0.676 $\pm$ 0.03 & 0.645 $\pm$ 0.03
& 0.637 $\pm$ 0.03 & 0.305 $\pm$ 0.03
& 0.586 $\pm$ 0.03 & 0.371 $\pm$ 0.03
& \textbf{0.764 $\pm$ 0.03} & 0.449 $\pm$ 0.04
& 0.803 $\pm$ 0.02 & 0.607 $\pm$ 0.03 \\

$\text{KLE}(K_\text{HEAT})$
& 0.775 $\pm$ 0.02 & 0.692 $\pm$ 0.03 
& 0.636 $\pm$ 0.03 & 0.311 $\pm$ 0.03
& 0.709 $\pm$ 0.03 & 0.452 $\pm$ 0.03
& 0.727 $\pm$ 0.03 & 0.448 $\pm$ 0.04
& 0.832 $\pm$ 0.02 & 0.633 $\pm$ 0.03 \\

EGKE(Ours)
& 0.776 $\pm$ 0.02 & \textbf{0.695 $\pm$ 0.03}
& 0.718 $\pm$ 0.03 & 0.344 $\pm$ 0.03
& \textbf{0.762 $\pm$ 0.03} & \textbf{0.484 $\pm$ 0.03}
& 0.727 $\pm$ 0.03 & 0.447 $\pm$ 0.04
& 0.865 $\pm$ 0.02 & 0.646 $\pm$ 0.03 \\
\bottomrule
\end{tabular}}
\end{table}

\begin{table}[H]
\centering
\caption{AUROC and AUARC on Falcon-7B-Instruct across datasets. Best results per metric in \textbf{bold}.}
\label{tab:falcon-7b-instruct}
\scriptsize
\resizebox{\textwidth}{!}{%
\begin{tabular}{lcccccccccc}
\toprule
\textbf{Method} &
\multicolumn{2}{c}{\textbf{TriviaQA}} &
\multicolumn{2}{c}{\textbf{SQuAD}} &
\multicolumn{2}{c}{\textbf{NQ}} &
\multicolumn{2}{c}{\textbf{SVAMP}} &
\multicolumn{2}{c}{\textbf{BioASQ}} \\
\cmidrule(lr){2-3}\cmidrule(lr){4-5}\cmidrule(lr){6-7}\cmidrule(lr){8-9}\cmidrule(lr){10-11}
 & AUROC & AUARC & AUROC & AUARC & AUROC & AUARC & AUROC & AUARC & AUROC & AUARC \\
\midrule
SE 
& \textbf{0.791 $\pm$ 0.02} & \textbf{0.549 $\pm$ 0.03} 
& 0.665 $\pm$ 0.04 & 0.251 $\pm$ 0.03 
& 0.716 $\pm$ 0.03 & 0.369 $\pm$ 0.03 
& 0.698 $\pm$ 0.04 & 0.279 $\pm$ 0.04 
& 0.831 $\pm$ 0.02 & 0.499 $\pm$ 0.03 \\

DSE
& 0.779 $\pm$ 0.02 & 0.532 $\pm$ 0.03 
& 0.655 $\pm$ 0.04 & 0.241 $\pm$ 0.03 
& 0.714 $\pm$ 0.03 & 0.351 $\pm$ 0.03 
& 0.711 $\pm$ 0.04 & 0.267 $\pm$ 0.03 
& 0.828 $\pm$ 0.02 & 0.479 $\pm$ 0.03 \\

PE
& 0.750 $\pm$ 0.03 & 0.522 $\pm$ 0.03
& 0.665 $\pm$ 0.04 & 0.257 $\pm$ 0.03
& 0.671 $\pm$ 0.03 & 0.348 $\pm$ 0.03
& \textbf{0.732 $\pm$ 0.04} & \textbf{0.290 $\pm$ 0.03}
& 0.655 $\pm$ 0.03 & 0.406 $\pm$ 0.03 \\

P\_true
& 0.544 $\pm$ 0.03 & 0.390 $\pm$ 0.03
& 0.404 $\pm$ 0.03 & 0.133 $\pm$ 0.02
& 0.554 $\pm$ 0.03 & 0.280 $\pm$ 0.03
& 0.592 $\pm$ 0.04 & 0.226 $\pm$ 0.03
& 0.598 $\pm$ 0.03 & 0.358 $\pm$ 0.03 \\

ER
& 0.719 $\pm$ 0.03 & 0.490 $\pm$ 0.03
& 0.628 $\pm$ 0.04 & 0.232 $\pm$ 0.03
& 0.658 $\pm$ 0.03 & 0.325 $\pm$ 0.03
& 0.683 $\pm$ 0.04 & 0.258 $\pm$ 0.03
& 0.829 $\pm$ 0.02 & 0.507 $\pm$ 0.03 \\

$\text{KLE}(K_\text{HEAT})$
& 0.771 $\pm$ 0.02 & 0.537 $\pm$ 0.03 
& 0.602 $\pm$ 0.04 & 0.223 $\pm$ 0.03
& 0.712 $\pm$ 0.03 & 0.362 $\pm$ 0.03
& 0.706 $\pm$ 0.04 & 0.281 $\pm$ 0.03
& 0.785 $\pm$ 0.02 & 0.468 $\pm$ 0.03 \\

EGKE(Ours)
& 0.782 $\pm$ 0.02 & 0.541 $\pm$ 0.03
& \textbf{0.672 $\pm$ 0.04} & \textbf{0.258 $\pm$ 0.03}
& \textbf{0.731 $\pm$ 0.03} & \textbf{0.371 $\pm$ 0.03}
& 0.661 $\pm$ 0.04 & 0.258 $\pm$ 0.03
& \textbf{0.845 $\pm$ 0.02} & \textbf{0.508 $\pm$ 0.03} \\
\bottomrule
\end{tabular}}
\end{table}

\begin{table}[H]
\centering
\caption{AUROC and AUARC on LLaMA-2-7B across datasets. Best results per column in \textbf{bold}.}
\label{tab:llama-2-7b}
\scriptsize
\resizebox{\textwidth}{!}{%
\begin{tabular}{lcccccccccc}
\toprule
\textbf{Method} &
\multicolumn{2}{c}{\textbf{TriviaQA}} &
\multicolumn{2}{c}{\textbf{SQuAD}} &
\multicolumn{2}{c}{\textbf{NQ}} &
\multicolumn{2}{c}{\textbf{SVAMP}} &
\multicolumn{2}{c}{\textbf{BioASQ}} \\
\cmidrule(lr){2-3}\cmidrule(lr){4-5}\cmidrule(lr){6-7}\cmidrule(lr){8-9}\cmidrule(lr){10-11}
 & AUROC & AUARC & AUROC & AUARC & AUROC & AUARC & AUROC & AUARC & AUROC & AUARC \\
\midrule
SE 
& 0.784 $\pm$ 0.03 & 0.745 $\pm$ 0.02 
& 0.785 $\pm$ 0.03 & 0.405 $\pm$ 0.03 
& 0.775 $\pm$ 0.02 & 0.541 $\pm$ 0.03 
& \textbf{0.878 $\pm$ 0.02} & \textbf{0.649 $\pm$ 0.03} 
& \textbf{0.852 $\pm$ 0.02} & \textbf{0.642 $\pm$ 0.03} \\

DSE
& 0.778 $\pm$ 0.02 & 0.742 $\pm$ 0.02 
& \textbf{0.798 $\pm$ 0.03} & 0.398 $\pm$ 0.03 
& 0.776 $\pm$ 0.02 & 0.527 $\pm$ 0.03 
& 0.870 $\pm$ 0.02 & 0.636 $\pm$ 0.03 
& 0.848 $\pm$ 0.02 & 0.618 $\pm$ 0.03 \\

PE
& 0.751 $\pm$ 0.03 & 0.730 $\pm$ 0.02
& 0.757 $\pm$ 0.03 & 0.408 $\pm$ 0.03
& 0.750 $\pm$ 0.03 & 0.533 $\pm$ 0.03
& 0.841 $\pm$ 0.02 & 0.632 $\pm$ 0.03
& 0.819 $\pm$ 0.02 & 0.620 $\pm$ 0.03 \\

P\_true
& 0.607 $\pm$ 0.03 & 0.668 $\pm$ 0.03
& 0.628 $\pm$ 0.03 & 0.331 $\pm$ 0.03
& 0.678 $\pm$ 0.03 & 0.480 $\pm$ 0.03
& 0.751 $\pm$ 0.03 & 0.555 $\pm$ 0.04
& 0.672 $\pm$ 0.03 & 0.521 $\pm$ 0.03 \\

ER
& 0.619 $\pm$ 0.03 & 0.664 $\pm$ 0.03
& 0.573 $\pm$ 0.03 & 0.288 $\pm$ 0.03
& 0.569 $\pm$ 0.03 & 0.403 $\pm$ 0.03
& 0.758 $\pm$ 0.03 & 0.574 $\pm$ 0.03
& 0.748 $\pm$ 0.02 & 0.583 $\pm$ 0.03 \\

$\text{KLE}(K_\text{HEAT})$
& 0.775 $\pm$ 0.03 & 0.743 $\pm$ 0.02 
& 0.710 $\pm$ 0.03 & 0.381 $\pm$ 0.03
& 0.780 $\pm$ 0.02 & 0.542 $\pm$ 0.03
& 0.867 $\pm$ 0.02 & 0.642 $\pm$ 0.03
& 0.818 $\pm$ 0.02 & 0.623 $\pm$ 0.03 \\

EGKE(Ours)
& \textbf{0.792 $\pm$ 0.02} & \textbf{0.755 $\pm$ 0.02}
& 0.784 $\pm$ 0.03 & \textbf{0.412 $\pm$ 0.03}
& \textbf{0.788 $\pm$ 0.02} & \textbf{0.545 $\pm$ 0.03}
& 0.856 $\pm$ 0.02 & 0.640 $\pm$ 0.03
& 0.850 $\pm$ 0.02 & 0.638 $\pm$ 0.03 \\
\bottomrule
\end{tabular}}
\end{table}
 
\begin{table}[H]
\centering
\caption{AUROC and AUARC on LLaMA-2-7B-Chat across datasets. Best results per column in \textbf{bold}.}
\label{tab:llama-2-7b-chat}
\scriptsize
\resizebox{\textwidth}{!}{%
\begin{tabular}{lcccccccccc}
\toprule
\textbf{Method} &
\multicolumn{2}{c}{\textbf{TriviaQA}} &
\multicolumn{2}{c}{\textbf{SQuAD}} &
\multicolumn{2}{c}{\textbf{NQ}} &
\multicolumn{2}{c}{\textbf{SVAMP}} &
\multicolumn{2}{c}{\textbf{BioASQ}} \\
\cmidrule(lr){2-3}\cmidrule(lr){4-5}\cmidrule(lr){6-7}\cmidrule(lr){8-9}\cmidrule(lr){10-11}
 & AUROC & AUARC & AUROC & AUARC & AUROC & AUARC & AUROC & AUARC & AUROC & AUARC \\
\midrule
SE 
& 0.771 $\pm$ 0.03 & 0.698 $\pm$ 0.02 
& 0.750 $\pm$ 0.03 & 0.360 $\pm$ 0.03 
& 0.724 $\pm$ 0.03 & \textbf{0.514 $\pm$ 0.03} 
& 0.841 $\pm$ 0.02 & 0.590 $\pm$ 0.03 
& \textbf{0.842 $\pm$ 0.02} & \textbf{0.640 $\pm$ 0.03} \\

DSE
& 0.787 $\pm$ 0.02 & 0.704 $\pm$ 0.02 
& 0.755 $\pm$ 0.03 & 0.356 $\pm$ 0.03 
& 0.726 $\pm$ 0.03 & 0.510 $\pm$ 0.03 
& \textbf{0.845 $\pm$ 0.02} & 0.588 $\pm$ 0.03 
& 0.840 $\pm$ 0.02 & 0.632 $\pm$ 0.03 \\

PE
& 0.737 $\pm$ 0.03 & 0.683 $\pm$ 0.03
& 0.719 $\pm$ 0.03 & 0.359 $\pm$ 0.03
& 0.715 $\pm$ 0.03 & 0.510 $\pm$ 0.03
& 0.835 $\pm$ 0.02 & 0.589 $\pm$ 0.03
& 0.678 $\pm$ 0.03 & 0.554 $\pm$ 0.03 \\

P\_true
& 0.694 $\pm$ 0.03 & 0.666 $\pm$ 0.03
& 0.598 $\pm$ 0.04 & 0.297 $\pm$ 0.03
& 0.647 $\pm$ 0.03 & 0.465 $\pm$ 0.03
& 0.573 $\pm$ 0.03 & 0.422 $\pm$ 0.04
& 0.790 $\pm$ 0.02 & 0.633 $\pm$ 0.03 \\

ER
& 0.643 $\pm$ 0.03 & 0.635 $\pm$ 0.03
& 0.562 $\pm$ 0.03 & 0.266 $\pm$ 0.03
& 0.570 $\pm$ 0.03 & 0.412 $\pm$ 0.03
& 0.803 $\pm$ 0.03 & 0.575 $\pm$ 0.03
& 0.729 $\pm$ 0.03 & 0.569 $\pm$ 0.03 \\

$\text{KLE}(K_\text{HEAT})$
& 0.791 $\pm$ 0.02 & 0.708 $\pm$ 0.03 
& 0.707 $\pm$ 0.03 & 0.349 $\pm$ 0.03
& 0.715 $\pm$ 0.03 & 0.508 $\pm$ 0.03
& \textbf{0.845 $\pm$ 0.02} & \textbf{0.591 $\pm$ 0.03}
& 0.805 $\pm$ 0.02 & 0.623 $\pm$ 0.03 \\

EGKE(Ours)
& \textbf{0.801 $\pm$ 0.02} & \textbf{0.717 $\pm$ 0.02}
& \textbf{0.757 $\pm$ 0.03} & \textbf{0.367 $\pm$ 0.03}
& \textbf{0.732 $\pm$ 0.02} & \textbf{0.514 $\pm$ 0.03}
& 0.834 $\pm$ 0.02 & 0.588 $\pm$ 0.03
& 0.838 $\pm$ 0.02 & 0.638 $\pm$ 0.03 \\
\bottomrule
\end{tabular}}
\end{table}

\begin{table}[H]
\centering
\caption{AUROC and AUARC on Mistral-7B-v0.1 across datasets. Best results per column in \textbf{bold}.}
\label{tab:mistral-7b-v0.1}
\scriptsize
\resizebox{\textwidth}{!}{%
\begin{tabular}{lcccccccccc}
\toprule
\textbf{Method} &
\multicolumn{2}{c}{\textbf{TriviaQA}} &
\multicolumn{2}{c}{\textbf{SQuAD}} &
\multicolumn{2}{c}{\textbf{NQ}} &
\multicolumn{2}{c}{\textbf{SVAMP}} &
\multicolumn{2}{c}{\textbf{BioASQ}} \\
\cmidrule(lr){2-3}\cmidrule(lr){4-5}\cmidrule(lr){6-7}\cmidrule(lr){8-9}\cmidrule(lr){10-11}
 & AUROC & AUARC & AUROC & AUARC & AUROC & AUARC & AUROC & AUARC & AUROC & AUARC \\
\midrule
SE 
& 0.771 $\pm$ 0.03 & 0.761 $\pm$ 0.02
& 0.750 $\pm$ 0.03 & 0.444 $\pm$ 0.03 
& \textbf{0.792 $\pm$ 0.02} & 0.607 $\pm$ 0.03 
& \textbf{0.884 $\pm$ 0.02} & \textbf{0.776 $\pm$ 0.02} 
& \textbf{0.870 $\pm$ 0.02} & \textbf{0.721 $\pm$ 0.02} \\

DSE
& 0.774 $\pm$ 0.02 & 0.760 $\pm$ 0.02 
& 0.758 $\pm$ 0.03 & 0.426 $\pm$ 0.03 
& 0.773 $\pm$ 0.02 & 0.598 $\pm$ 0.03 
& 0.867 $\pm$ 0.02 & 0.761 $\pm$ 0.02 
& 0.861 $\pm$ 0.02 & 0.707 $\pm$ 0.02 \\

PE
& 0.766 $\pm$ 0.02 & \textbf{0.766 $\pm$ 0.02}
& 0.749 $\pm$ 0.03 & \textbf{0.452 $\pm$ 0.03}
& 0.740 $\pm$ 0.02 & 0.580 $\pm$ 0.03
& 0.824 $\pm$ 0.02 & 0.756 $\pm$ 0.02
& 0.742 $\pm$ 0.02 & 0.653 $\pm$ 0.03 \\

P\_true
& 0.714 $\pm$ 0.03 & 0.731 $\pm$ 0.03
& 0.715 $\pm$ 0.03 & 0.424 $\pm$ 0.03
& 0.715 $\pm$ 0.02 & 0.563 $\pm$ 0.03
& 0.745 $\pm$ 0.03 & 0.701 $\pm$ 0.03
& 0.734 $\pm$ 0.03 & 0.658 $\pm$ 0.03 \\

ER
& 0.642 $\pm$ 0.03 & 0.698 $\pm$ 0.03
& 0.640 $\pm$ 0.03 & 0.371 $\pm$ 0.03
& 0.581 $\pm$ 0.03 & 0.466 $\pm$ 0.03
& 0.853 $\pm$ 0.02 & 0.769 $\pm$ 0.02
& 0.773 $\pm$ 0.02 & 0.674 $\pm$ 0.03 \\

$\text{KLE}(K_\text{HEAT})$
& 0.782 $\pm$ 0.02 & \textbf{0.766 $\pm$ 0.02}
& 0.708 $\pm$ 0.03 & 0.430 $\pm$ 0.03
& 0.772 $\pm$ 0.02 & 0.559 $\pm$ 0.03
& 0.877 $\pm$ 0.02 & 0.774 $\pm$ 0.02
& 0.813 $\pm$ 0.02 & 0.692 $\pm$ 0.02 \\

EGKE(Ours)
& \textbf{0.789 $\pm$ 0.02} & \textbf{0.766 $\pm$ 0.02}
& \textbf{0.770 $\pm$ 0.02} & \textbf{0.452 $\pm$ 0.03}
& 0.788 $\pm$ 0.02 & \textbf{0.608 $\pm$ 0.03}
& 0.849 $\pm$ 0.02 & 0.763 $\pm$ 0.02
& 0.866 $\pm$ 0.02 & 0.715 $\pm$ 0.02 \\

\bottomrule
\end{tabular}}
\end{table}

\begin{table}[H]
\centering
\caption{AUROC and AUARC on Mistral-7B-v0.1-instruct across datasets. Best results per column in \textbf{bold}.}
\label{tab:mistral-7b-v0.1-instruct}
\scriptsize
\resizebox{\textwidth}{!}{%
\begin{tabular}{lcccccccccc}
\toprule
\textbf{Method} &
\multicolumn{2}{c}{\textbf{TriviaQA}} &
\multicolumn{2}{c}{\textbf{SQuAD}} &
\multicolumn{2}{c}{\textbf{NQ}} &
\multicolumn{2}{c}{\textbf{SVAMP}} &
\multicolumn{2}{c}{\textbf{BioASQ}} \\
\cmidrule(lr){2-3}\cmidrule(lr){4-5}\cmidrule(lr){6-7}\cmidrule(lr){8-9}\cmidrule(lr){10-11}
 & AUROC & AUARC & AUROC & AUARC & AUROC & AUARC & AUROC & AUARC & AUROC & AUARC \\
\midrule
SE 
& 0.803 $\pm$ 0.02 & \textbf{0.676 $\pm$ 0.03}
& 0.756 $\pm$ 0.03 & 0.367 $\pm$ 0.03 
& 0.774 $\pm$ 0.03 & 0.457 $\pm$ 0.03 
& 0.898 $\pm$ 0.02 & 0.730 $\pm$ 0.03 
& \textbf{0.901 $\pm$ 0.02} & \textbf{0.666 $\pm$ 0.03} \\

DSE
& \textbf{0.811 $\pm$ 0.02} & \textbf{0.676 $\pm$ 0.03} 
& 0.760 $\pm$ 0.03 & 0.361 $\pm$ 0.03 
& 0.781 $\pm$ 0.02 & 0.441 $\pm$ 0.03 
& 0.895 $\pm$ 0.02 & 0.721 $\pm$ 0.03 
& 0.896 $\pm$ 0.02 & 0.645 $\pm$ 0.02 \\

PE
& 0.784 $\pm$ 0.02 & 0.662 $\pm$ 0.03
& 0.748 $\pm$ 0.03 & 0.374 $\pm$ 0.03
& 0.740 $\pm$ 0.03 & 0.434 $\pm$ 0.03
& 0.854 $\pm$ 0.02 & 0.713 $\pm$ 0.03
& 0.740 $\pm$ 0.03 & 0.570 $\pm$ 0.03 \\

P\_true
& 0.742 $\pm$ 0.02 & 0.632 $\pm$ 0.03
& 0.608 $\pm$ 0.03 & 0.292 $\pm$ 0.03
& 0.706 $\pm$ 0.03 & 0.419 $\pm$ 0.03
& 0.760 $\pm$ 0.03 & 0.658 $\pm$ 0.03
& 0.736 $\pm$ 0.03 & 0.578 $\pm$ 0.03 \\

ER
& 0.677 $\pm$ 0.03 & 0.605 $\pm$ 0.03
& 0.620 $\pm$ 0.03 & 0.301 $\pm$ 0.03
& 0.598 $\pm$ 0.03 & 0.333 $\pm$ 0.03
& 0.860 $\pm$ 0.02 & 0.710 $\pm$ 0.03
& 0.805 $\pm$ 0.02 & 0.605 $\pm$ 0.03 \\

$\text{KLE}(K_\text{HEAT})$
& 0.797 $\pm$ 0.02 & \textbf{0.676 $\pm$ 0.03} 
& 0.680 $\pm$ 0.03 & 0.338 $\pm$ 0.03
& 0.747 $\pm$ 0.03 & 0.443 $\pm$ 0.03
& \textbf{0.899 $\pm$ 0.02} & \textbf{0.732 $\pm$ 0.02}
& 0.851 $\pm$ 0.02 & 0.641 $\pm$ 0.03 \\

EGKE(Ours)
& 0.806 $\pm$ 0.02 & \textbf{0.676 $\pm$ 0.03}
& \textbf{0.780 $\pm$ 0.03} & \textbf{0.380 $\pm$ 0.03}
& \textbf{0.794 $\pm$ 0.02} & \textbf{0.465 $\pm$ 0.03}
& 0.885 $\pm$ 0.02 & 0.727 $\pm$ 0.03
& 0.898 $\pm$ 0.02 & 0.659 $\pm$ 0.03 \\
\bottomrule
\end{tabular}}
\end{table}



\end{document}